\documentclass[11pt]{article}

\usepackage[utf8]{inputenc}
\usepackage[T1]{fontenc}
\usepackage[margin=1in]{geometry}

\usepackage[numbers,compress]{natbib}

\usepackage{graphicx}
\usepackage{booktabs}
\usepackage{amsmath}
\usepackage{amssymb}
\usepackage{amsfonts}
\usepackage{nicefrac}
\usepackage{microtype}
\usepackage{xcolor}
\usepackage{url}
\usepackage[colorlinks=true,
            linkcolor=black,
            citecolor=blue,
            urlcolor=blue]{hyperref}

\title{
\textbf{State-Aware Interaction MIL for Rare Joint Molecular Phenotype Prediction in Colorectal Cancer and Lung Adenocarcinoma}%
\thanks{\textcolor{blue}{\textbf{Accepted at the NeurIPS 2026 Workshop on AI at Scale for Clinical Impact (ASCI): Cancer Pathology Foundation Models.}}}
}

\author{
\textbf{Dasari Naga Raju}\textsuperscript{1 *}
\hspace{0.001em}
\textbf{Tripti Bameta}\textsuperscript{2}
\\[0.5em]
\small \textsuperscript{1}Independent Researcher, Andhra Pradesh, India
\\
\small \textsuperscript{2}Tata Memorial Centre-ACTREC, Navi Mumbai, India
\\[0.4em]
\small \textsuperscript{*}Corresponding Author:
\href{mailto:raajuuu1998@gmail.com}{raajuuu1998@gmail.com}
}

\date{}

\begin{document}

\maketitle

\begin{abstract}
Joint molecular phenotype prediction is complicated by small joint-positive
populations and overlapping histological features across alternative
molecular states. Existing computational pathology approaches typically
predict biomarkers independently or formulate the joint-positive phenotype
as a binary endpoint. Independent prediction does not model interactions
between biomarker-specific histological representations, whereas binary
joint prediction collapses the double-negative and two single-positive
configurations into a single negative class. We propose
\textbf{State-Aware Interaction MIL}, a weakly supervised method that
preserves biomarker-specific histological representations, models their
interaction, and supervises the complete four-state molecular configuration.
We evaluate the proposed approach for joint BRAF$^{+}$/MSI$^{+}$ prediction
in colorectal cancer and EGFR$^{+}$/TP53$^{+}$ prediction in lung
adenocarcinoma using frozen UNI2-h and CONCH pathology foundation-model
representations. With UNI2-h, State-Aware Interaction MIL achieved an average precision of
$0.5566$ in colorectal cancer (joint-positive prevalence 6.8\%) compared
with $0.5161$ for NaiveMTL, and $0.2784$ in lung adenocarcinoma
(joint-positive prevalence 8.6\%) compared with $0.2525$ for
IndependentPair. With CONCH, State-Aware achieved an average precision of
$0.4410$ compared with $0.3932$ for DirectJoint in colorectal cancer and
$0.1659$ compared with $0.1226$ for DirectJoint in lung adenocarcinoma. Four-state analysis
showed that joint-positive prediction scores differed across molecular
configurations, and the joint-positive group had elevated median scores in
both cohorts. These results indicate that pathology
foundation-model representations contain predictive information for rare
joint molecular phenotypes and that preserving biomarker-specific
representations within a structured molecular-state formulation can improve
prediction of these phenotypes from histopathology.
\end{abstract}

\section{Introduction}

Joint molecular phenotypes can provide information beyond individual
biomarkers when biomarker co-occurrence defines distinct disease subgroups
or influences treatment response. In colorectal cancer, microsatellite
instability~(MSI) and BRAF mutation are clinically relevant molecular
biomarkers with therapeutic and prognostic implications. MSI occurs in
approximately 10--18\% of localized colorectal cancers but in about 5\% of
metastatic disease and is a predictive biomarker for response to
immune-checkpoint inhibition~\citep{cohen2021msi,andre2020pembrolizumab}.
BRAF mutation is detected in approximately 8--12\% of metastatic colorectal
cancers and is associated with poor prognosis; BRAF status also informs
molecular stratification and treatment planning~\citep{nakayama2020braf}. BRAF
mutation and MSI are also associated with characteristic clinicopathological
features, including proximal tumor location, mucinous or poorly
differentiated morphology, and prominent immune-cell infiltration
~\citep{li2006,rosenbaum2016,smyrk2001,venderbosch2014}. Within MSI-positive
colorectal cancer, BRAF is strongly associated with sporadic
MLH1-methylated disease and is uncommon in Lynch syndrome-associated CRC;
combined BRAF/MSI assessment therefore contributes to molecular
classification and hereditary-risk evaluation~\citep{parsons2012braf}.
Combined BRAF/MSI status has also been associated with distinct prognostic
patterns~\citep{yang2018brafmsi}. In lung adenocarcinoma, TP53 co-mutations
are frequent in EGFR-mutant tumors and have been associated with shorter
progression-free and overall survival during EGFR-targeted therapy,
supporting the clinical relevance of EGFR/TP53
co-occurrence~\citep{liu2022egfrtp53,lan2022tp53}. These clinical distinctions motivate joint molecular phenotype prediction
from histopathology.

Routine H\&E whole-slide images contain morphological patterns associated
with molecular biomarker status, enabling computational prediction from
histopathology. MSI has been predicted from gastrointestinal cancer
histology~\citep{kather2019}, while EGFR, TP53, and other recurrent
molecular biomarkers have been predicted from lung adenocarcinoma
whole-slide images~\citep{coudray2018,tomita2022}. Because molecular labels
are generally available at the patient or slide level rather than for
individual tissue regions, multiple-instance learning~(MIL) provides a
weakly supervised framework for learning slide-level predictions from
collections of unlabeled tissue instances. Pathology foundation models such
as UNI2-h and CONCH provide pretrained tile representations for downstream
whole-slide analysis~\citep{chen2024uni,uni2h,lu2024conch}.
Attention-based MIL introduced learnable instance weighting for slide-level
aggregation~\citep{ilse2018}; CLAM extends this approach with instance-level
clustering constraints~\citep{lu2021clam}, while TransMIL models
dependencies among tissue instances using transformer-based
aggregation~\citep{shao2021transmil}. These approaches provide several
mechanisms for aggregating heterogeneous tissue information into slide-level
representations, but they were not designed specifically for structured
joint molecular-state prediction. Most histology-based molecular prediction studies treat biomarkers such as
MSI, BRAF, EGFR, and TP53 as separate prediction endpoints
~\citep{kather2019,niehues2023,wagner2023,coudray2018,tomita2022}.
Multi-task learning jointly optimizes several prediction objectives, often
through shared representations, while multi-label methods can additionally
model dependencies among multiple labels~\citep{caruana1997,read2011}.
Clinically relevant biomarker co-occurrence has received limited attention
in computational pathology, particularly for rare joint-positive molecular
phenotypes.

We propose \textbf{State-Aware Interaction MIL}, a weakly supervised MIL
method for rare joint molecular phenotype prediction from whole-slide
histology. The main contributions of this work are:

\begin{itemize}

\item We preserve biomarker-specific histological representations through
separate MIL branches and model interactions between the resulting slide
representations.

\item We supervise the complete four-state molecular configuration,
distinguishing the double-negative, two single-positive, and joint-positive
states rather than collapsing the three non-joint-positive states into a
single negative class.
\end{itemize}

\section{Method}
\subsection{State-Aware Interaction MIL}

\begin{figure*}[!htb]
    \centering
    \includegraphics[width=\textwidth]{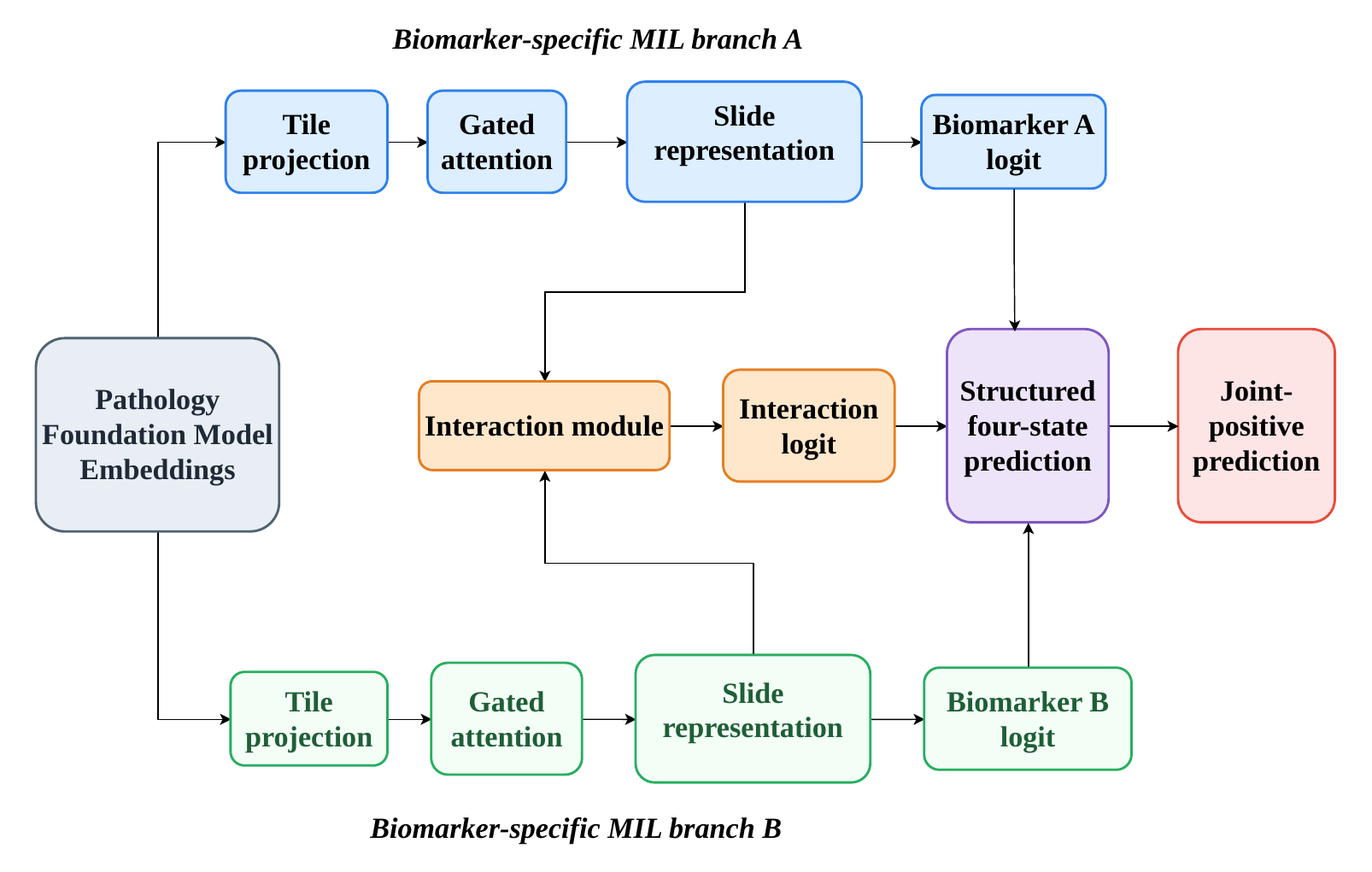}
\caption{Architecture of State-Aware Interaction MIL, comprising
biomarker-specific MIL branches, an interaction module, and a structured
four-state molecular prediction head.}
    \label{fig:stateaware_architecture}
\end{figure*}

\paragraph{Problem formulation.}
Let a patient be represented by a bag of frozen pathology foundation-model
tile embeddings
\begin{equation}
\mathbf{X}=\{\mathbf{x}_1,\ldots,\mathbf{x}_N\},
\qquad
\mathbf{x}_i\in\mathbb{R}^{D},
\end{equation}
where $D=1536$ for UNI2-h and $D=512$ for CONCH. Let
$A,B\in\{0,1\}$ denote binary indicators for two molecular biomarkers.
The corresponding molecular-state class index is defined as
\begin{equation}
y_s=2A+B\in\{0,1,2,3\}.
\end{equation}
The four biomarker configurations $00$, $01$, $10$, and $11$ therefore
correspond to class indices $0$, $1$, $2$, and $3$, respectively. State
$00$ denotes both biomarkers negative, $01$ denotes biomarker $B$ positive
only, $10$ denotes biomarker $A$ positive only, and $11$ denotes the
joint-positive phenotype in which both biomarkers are positive. State $11$
serves as the primary prediction target. For colorectal cancer, we define
$(A,B)=(\mathrm{BRAF},\mathrm{MSI})$, whereas for lung adenocarcinoma,
$(A,B)=(\mathrm{EGFR},\mathrm{TP53})$.
Figure~\ref{fig:stateaware_architecture} illustrates the
biomarker-specific MIL branches, interaction module, and structured
four-state prediction head.

\paragraph{Biomarker-specific MIL representations.}
State-Aware Interaction MIL uses two separately parameterized gated-attention
MIL branches~\citep{ilse2018}, each with its own tile projection and attention distribution,
to produce biomarker-specific slide representations for the corresponding
molecular targets. For biomarker $k\in\{A,B\}$, tile embeddings are
projected to 256-dimensional features
$\mathbf{h}_{k,i}\in\mathbb{R}^{256}$ and assigned normalized
gated-attention weights $\alpha_{k,i}$. The corresponding slide-level
representation is obtained as
\begin{equation}
\mathbf{z}_k
=
\sum_{i=1}^{N}
\alpha_{k,i}\mathbf{h}_{k,i},
\qquad
\sum_{i=1}^{N}\alpha_{k,i}=1,
\end{equation}
where the gated-attention module uses an attention dimension of 128.
The two branch representations are therefore
\begin{equation}
\mathbf{z}_A=\mathrm{MIL}_A(\mathbf{X}),
\qquad
\mathbf{z}_B=\mathrm{MIL}_B(\mathbf{X}),
\qquad
\mathbf{z}_A,\mathbf{z}_B\in\mathbb{R}^{256}.
\end{equation}
Biomarker-specific linear classifiers map these representations to scalar
logits,
\begin{equation}
a=f_A(\mathbf{z}_A),
\qquad
b=f_B(\mathbf{z}_B).
\end{equation}

\paragraph{Explicit biomarker interaction.}
Interactions between the biomarker-specific slide representations are modeled
through the interaction vector
\begin{equation}
\mathbf{u}
=
\left[
\mathbf{z}_A,\,
\mathbf{z}_B,\,
\mathbf{z}_A\odot\mathbf{z}_B,\,
|\mathbf{z}_A-\mathbf{z}_B|
\right]
\in\mathbb{R}^{1024}.
\label{eq:interaction}
\end{equation}
The interaction vector retains both biomarker-specific representations and
augments them with their elementwise product and absolute difference.
An interaction MLP
\begin{equation}
g:\mathbb{R}^{1024}
\rightarrow
\mathbb{R}^{256}
\rightarrow
\mathbb{R}
\end{equation}
maps $\mathbf{u}$ to a scalar interaction logit,
\begin{equation}
i=g(\mathbf{u}).
\end{equation}

\paragraph{Structured four-state prediction.}
The four molecular-state logits are defined through a structured
parameterization of the two biomarker-specific logits and their learned
interaction:
\begin{equation}
\mathbf{q}
=
[q_{00},q_{01},q_{10},q_{11}]
=
[0,\;b,\;a,\;a+b+i].
\label{eq:state_logits}
\end{equation}
The double-negative state serves as the reference category and is assigned
$q_{00}=0$. The two single-positive states are parameterized by their
corresponding biomarker-specific logits, whereas the joint-positive state
combines both biomarker-specific logits with the learned interaction term.

When \(i=0\), the joint-positive logit reduces to the additive combination
of the two biomarker-specific logits,
\begin{equation}
q_{11}=a+b.
\end{equation}

A nonzero $i$ allows the model to capture departures from the additive
combination of the two biomarker-specific logits when estimating the joint
molecular state.

Molecular-state probabilities are obtained by
\begin{equation}
\mathbf{p}
=
\operatorname{softmax}(\mathbf{q})
=
[p_{00},p_{01},p_{10},p_{11}],
\end{equation}
and the final joint-positive prediction is
\begin{equation}
p_{\mathrm{joint}}
=
P(A=1,B=1\mid\mathbf{X})
=
p_{11}.
\label{eq:joint_prediction}
\end{equation}

\paragraph{Training objective.}
The model is trained using class-weighted cross-entropy over the four
molecular states. For a patient with state label $y_s$, the state loss is
\begin{equation}
\mathcal{L}_{\mathrm{state}}
=
-w_{y_s}\log p_{y_s},
\end{equation}
where the class weight for state $c$ is
\begin{equation}
w_c
=
\frac{N_{\mathrm{train}}}{4N_c}.
\end{equation}
Here, $N_{\mathrm{train}}$ denotes the number of patients in the training
partition and $N_c$ denotes the number assigned to state $c$.

To provide direct biomarker-specific supervision to the two MIL branches,
weighted binary cross-entropy losses $\mathcal{L}_A$ and $\mathcal{L}_B$
are applied to logits $a$ and $b$, respectively. The corresponding
positive-class weights are
\begin{equation}
w_A^{+}
=
\frac{N_A^{-}}{N_A^{+}},
\qquad
w_B^{+}
=
\frac{N_B^{-}}{N_B^{+}},
\end{equation}
where $N_k^{+}$ and $N_k^{-}$ denote the numbers of positive and negative
training patients for biomarker $k$, respectively. The overall objective is
\begin{equation}
\mathcal{L}
=
\mathcal{L}_{\mathrm{state}}
+
\lambda_{\mathrm{aux}}
\frac{\mathcal{L}_{A}+\mathcal{L}_{B}}{2},
\qquad
\lambda_{\mathrm{aux}}=0.25.
\label{eq:loss}
\end{equation}

\section{Experimental Setup}

\subsection{Dataset and Labels}

State-Aware Interaction MIL and the comparison models are evaluated on two
TCGA cohorts for distinct joint molecular phenotype prediction tasks. The
colorectal cancer cohort comprises 443 patients from
TCGA-COAD+READ~\citep{tcga_crc} with both BRAF and MSI labels, and the lung
adenocarcinoma cohort comprises 428 patients from
TCGA-LUAD~\citep{tcga_luad} with both EGFR and TP53 labels. Only patients with
valid labels for both biomarkers and available foundation-model
representations were retained. The primary
prediction endpoint is the joint-positive phenotype, defined as
BRAF$^{+}$/MSI$^{+}$ in colorectal cancer and
EGFR$^{+}$/TP53$^{+}$ in lung adenocarcinoma. Cohort composition and
joint-positive prevalence are reported in
Table~\ref{tab:dataset_composition}.

\begin{table}[htb]
\caption{Cohort composition for joint molecular phenotype prediction.
States $00$, $01$, $10$, and $11$ denote the four biomarker configurations,
with state $11$ representing the joint-positive phenotype.}
\label{tab:dataset_composition}
\centering
\small
\setlength{\tabcolsep}{3.5pt}
\resizebox{\linewidth}{!}{
\begin{tabular}{cccccccc}
\toprule
\textbf{Cohort} &
\textbf{Biomarker pair} &
\textbf{$N$} &
\textbf{$00$} &
\textbf{$01$} &
\textbf{$10$} &
\textbf{$11$} &
\textbf{Joint prevalence} \\
\midrule
TCGA-COAD+READ &
BRAF/MSI &
443 &
375 &
27 &
11 &
30 &
6.8\% \\
TCGA-LUAD &
EGFR/TP53 &
428 &
186 &
184 &
21 &
37 &
8.6\% \\
\bottomrule
\end{tabular}}
\end{table}

\subsection{Foundation Models and Baselines}

State-Aware Interaction MIL and all comparison models use frozen
representations from UNI2-h~\citep{chen2024uni,uni2h} and
CONCH~\citep{lu2024conch}, which produce 1,536- and 512-dimensional tile
embeddings, respectively. Whole-slide images are processed at $20\times$
magnification ($0.5~\mu\mathrm{m}$/pixel), with tissue regions identified
using Otsu thresholding~\citep{otsu1979} and divided into
$224\times224$-pixel tiles. During training, up to 3,000 tissue tiles are
randomly sampled per patient, while all available tile embeddings are used
for validation and testing. \textit{DirectJoint} uses a single gated-attention MIL
model~\citep{ilse2018} to distinguish the joint-positive phenotype from the
remaining molecular configurations. \textit{IndependentPair} trains two
separate biomarker-specific MIL models and derives the joint prediction as
the product of their marginal probabilities. \textit{NaiveMTL} uses a shared
gated-attention slide representation with separate prediction heads for
biomarker $A$, biomarker $B$, and the joint-positive endpoint, without
explicitly modeling interactions between biomarker-specific representations.
\textit{PostHoc-LR} combines the logits of the independently trained
biomarker models using logistic regression. FourState-ABMIL uses a single
gated-attention MIL encoder followed by a four-class linear head trained on
the 00/01/10/11 molecular states using the same class-weighted
cross-entropy used for state supervision in State-Aware.

\subsection{Metrics and Training Parameters}

Joint-positive prediction performance is evaluated using area under the
receiver operating characteristic curve~(AUROC) and average precision~(AP).
Results are reported as mean~$\pm$ standard deviation across five
patient-level outer folds stratified by the four-state molecular label.
The TCGA-COAD+READ test folds contain 83--93 patients with 5--7
joint-positive cases, while the TCGA-LUAD test folds contain 85--86 patients
with 7--8 joint-positive cases. Within each outer-training partition, 15\%
of patients are reserved for inner validation, with stratification based on
the four-state molecular label. All neural network models are trained using AdamW~\citep{loshchilov2019} with a learning rate of
$10^{-4}$, weight decay of $10^{-5}$, and dropout of $0.10$ for a maximum
of 30 epochs, with gradient accumulation over four patient bags. Checkpoints
are selected by validation AP, with early stopping after six epochs without
improvement.
For \textit{PostHoc-LR}, biomarker logits are standardized using
training-set statistics and combined with class-balanced logistic regression
using $C=1.0$, the \texttt{liblinear} solver, and a maximum of 2,000
iterations.

\section{Results}
\subsection{Joint-Positive Prediction Performance}
Table~\ref{tab:main_results} summarizes joint-positive prediction
performance across both cohorts and foundation-model representations.
With UNI2-h, State-Aware Interaction MIL achieved AP values of $0.5566$
in TCGA-COAD+READ and $0.2784$ in TCGA-LUAD, compared with $0.5161$
for NaiveMTL and $0.2525$ for IndependentPair, corresponding to
improvements of $0.0405$ and $0.0259$, respectively. State-Aware AUROC was lower than NaiveMTL in TCGA-COAD+READ and PostHoc-LR
in TCGA-LUAD.
\begin{table*}[htb]
\caption{Joint-positive molecular phenotype prediction for
BRAF$^{+}$/MSI$^{+}$ in TCGA-COAD+READ (prevalence 6.8\%) and
EGFR$^{+}$/TP53$^{+}$ in TCGA-LUAD (prevalence 8.6\%) using UNI2-h and
CONCH pathology foundation-model representations.}
\label{tab:main_results}
\centering
\footnotesize
\setlength{\tabcolsep}{4pt}
\renewcommand{\arraystretch}{0.95}
\begin{tabular}{@{}clcccc@{}}
\toprule
& &
\multicolumn{2}{c}{\textbf{TCGA-COAD+READ}} &
\multicolumn{2}{c}{\textbf{TCGA-LUAD}} \\
\cmidrule(lr){3-4}
\cmidrule(lr){5-6}
\textbf{FM} & \textbf{Method} &
\textbf{AUROC} & \textbf{AP} &
\textbf{AUROC} & \textbf{AP} \\
\midrule

\textbf{UNI2-h}
& DirectJoint
& $0.8998\pm0.0580$ & $0.4489\pm0.1479$
& $0.6264\pm0.1122$ & $0.1928\pm0.1262$ \\

&
IndependentPair
& $0.9090\pm0.0726$ & $0.4864\pm0.1146$
& $0.6842\pm0.1476$ & $0.2525\pm0.1347$ \\

&
NaiveMTL
& $\mathbf{0.9096\pm0.0627}$ & $0.5161\pm0.1531$
& $0.6485\pm0.1724$ & $0.2202\pm0.1820$ \\

&
PostHoc-LR
& $0.9093\pm0.0776$ & $0.4907\pm0.1122$
& $\mathbf{0.6894\pm0.1459}$ & $0.2177\pm0.1224$ \\

&
\textbf{State-Aware}
& $0.9080\pm0.0965$ & $\mathbf{0.5566\pm0.2015}$
& $0.6725\pm0.1746$ & $\mathbf{0.2784\pm0.1714}$ \\

\midrule

\textbf{CONCH}
& DirectJoint
& $0.8780\pm0.0565$ & $0.3932\pm0.0658$
& $0.5292\pm0.1481$ & $0.1226\pm0.0728$ \\

&
IndependentPair
& $0.8372\pm0.0605$ & $0.3448\pm0.0916$
& $0.4926\pm0.0869$ & $0.1012\pm0.0178$ \\

&
NaiveMTL
& $0.8269\pm0.0317$ & $0.3225\pm0.0502$
& $0.4996\pm0.1063$ & $0.1101\pm0.0345$ \\

&
PostHoc-LR
& $0.8441\pm0.0622$ & $0.3430\pm0.1020$
& $0.4990\pm0.1033$ & $0.1049\pm0.0219$ \\

&
\textbf{State-Aware}
& $\mathbf{0.8876\pm0.0473}$ & $\mathbf{0.4410\pm0.0446}$
& $\mathbf{0.5725\pm0.1285}$ & $\mathbf{0.1659\pm0.1078}$ \\

\bottomrule
\end{tabular}
\end{table*}
In TCGA-COAD+READ, AP increased from $0.3932$ with DirectJoint
to $0.4410$, while AUROC increased from $0.8780$ to $0.8876$. In
TCGA-LUAD, AP increased from $0.1226$ with DirectJoint to $0.1659$, and
AUROC increased from $0.5292$ to $0.5725$. The joint-positive prevalence was 6.8\% in TCGA-COAD+READ and 8.6\% in TCGA-LUAD, corresponding to prevalence-based reference AP values of approximately 0.068 and 0.086, respectively.

\begin{figure*}[htb]
    \centering

    \begin{minipage}{0.48\textwidth}
        \centering
        \includegraphics[width=\linewidth]{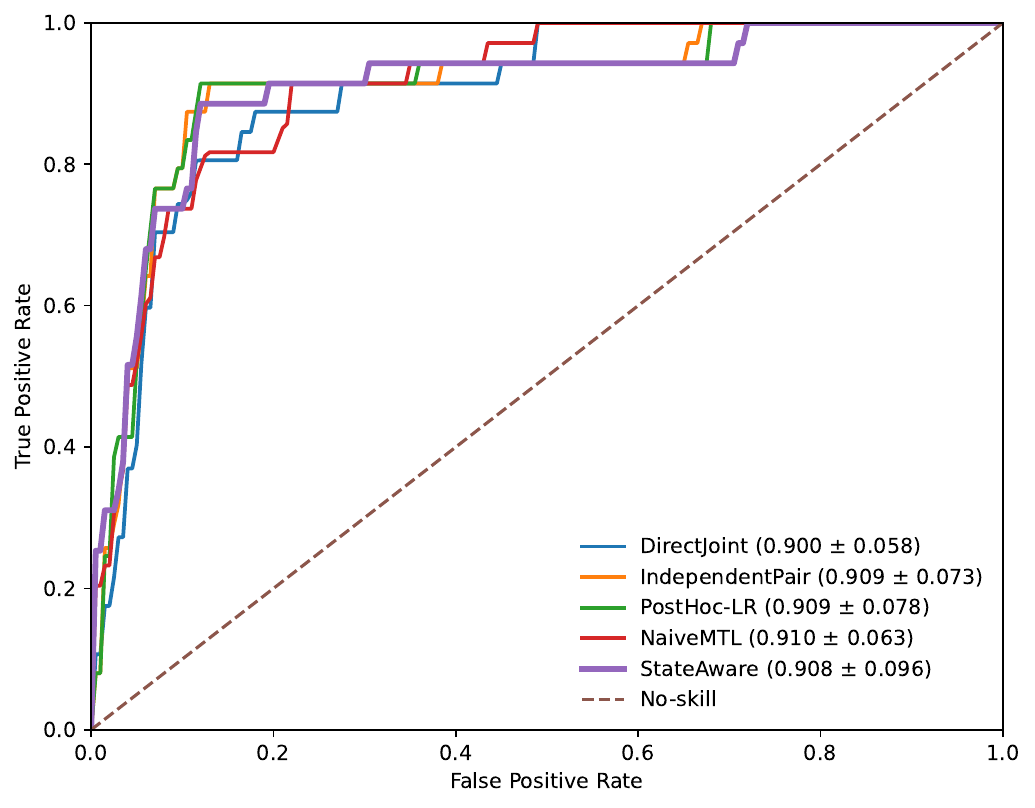}
        \textbf{(a) ROC, TCGA-COAD+READ}
    \end{minipage}
    \hfill
    \begin{minipage}{0.48\textwidth}
        \centering
        \includegraphics[width=\linewidth]{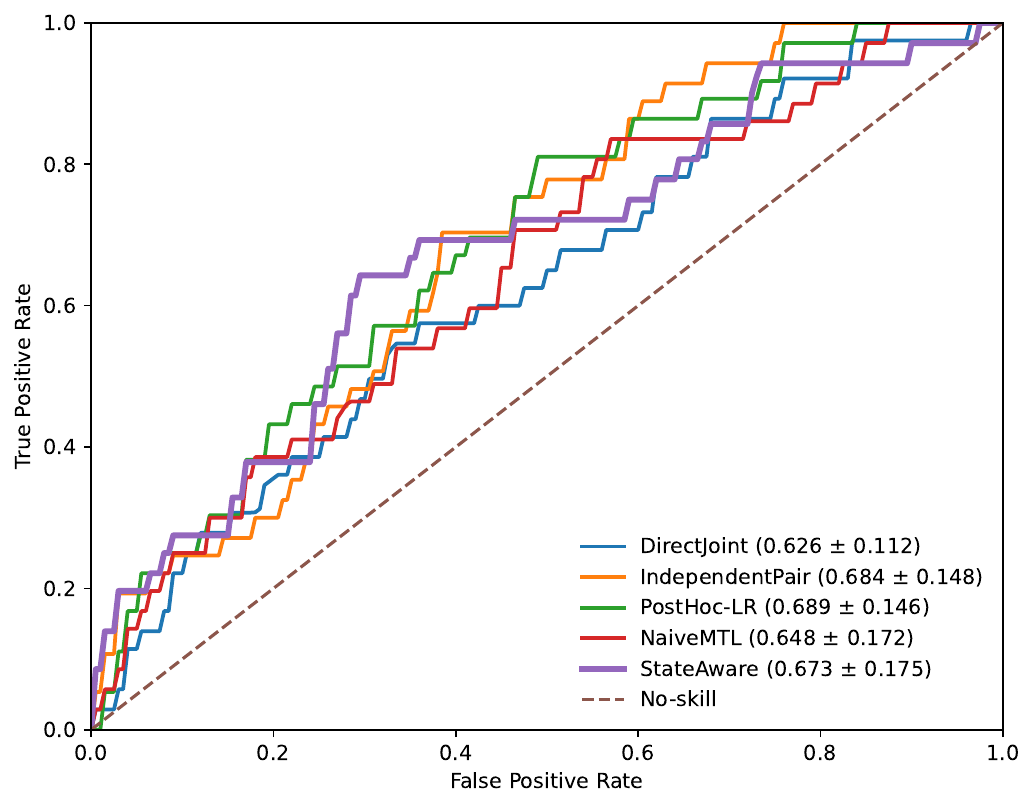}
        \textbf{(b) ROC, TCGA-LUAD}
    \end{minipage}

    \vspace{0.6em}

    \begin{minipage}{0.48\textwidth}
        \centering
        \includegraphics[width=\linewidth]{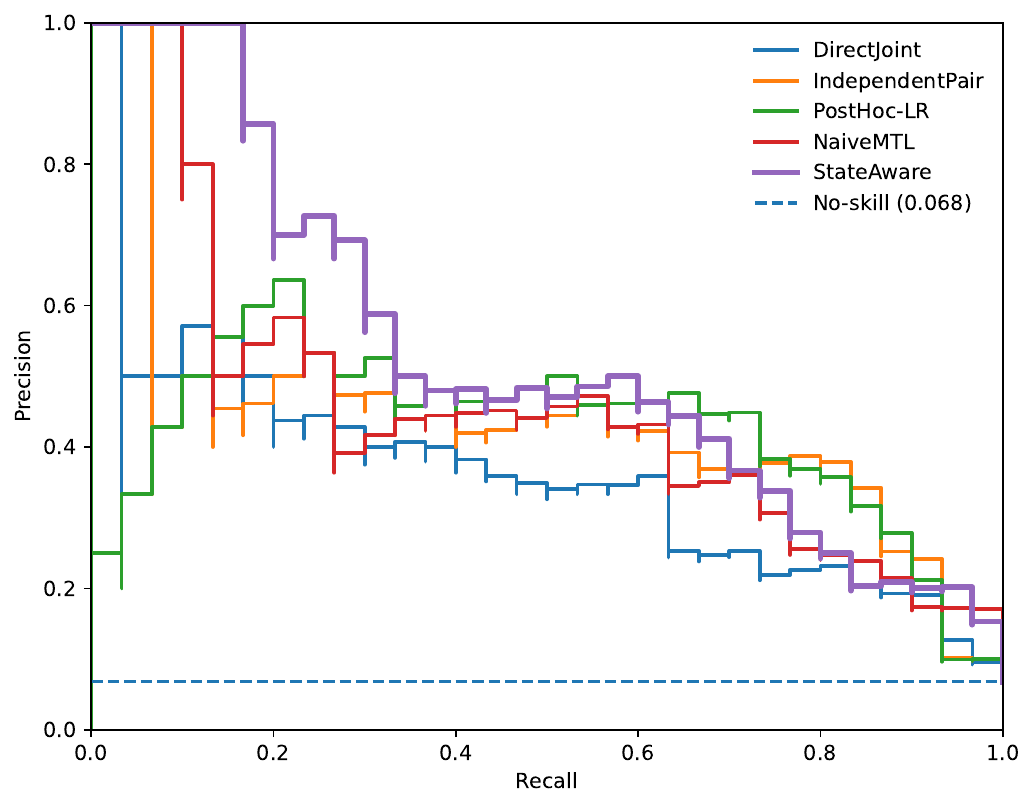}
        \textbf{(c) Precision--recall, TCGA-COAD+READ}
    \end{minipage}
    \hfill
    \begin{minipage}{0.48\textwidth}
        \centering
        \includegraphics[width=\linewidth]{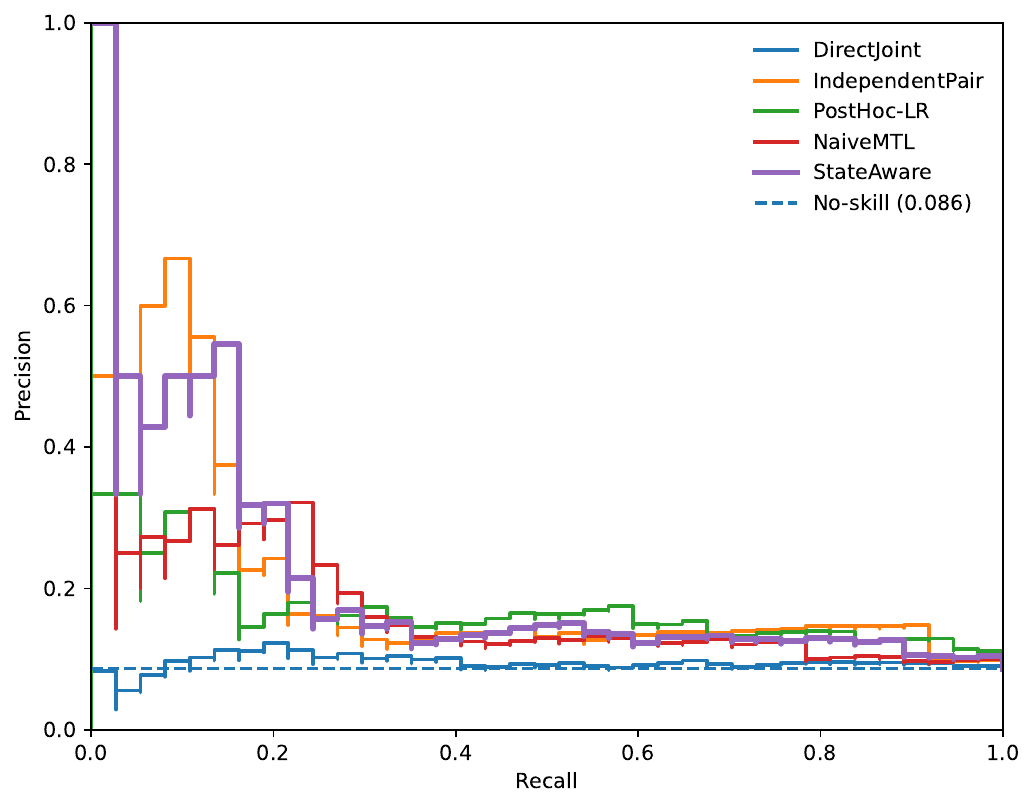}
        \textbf{(d) Precision--recall, TCGA-LUAD}
    \end{minipage}

\caption{ROC and precision--recall curves for UNI2-h-based joint-positive
molecular phenotype prediction in TCGA-COAD+READ and TCGA-LUAD.}
    \label{fig:roc_pr_curves}
\end{figure*}

Across both foundation-model representations, State-Aware improved AP
relative to the comparison methods in both cohorts, indicating that the AP
improvement was not specific to a single pathology representation. The
AUROC results differed by representation: State-Aware was below the leading
comparison method in both UNI2-h experiments, while its AUROC was above the
comparison methods in both CONCH experiments. The corresponding UNI2-h ROC and precision--recall curves are shown in
Figure~\ref{fig:roc_pr_curves}. In TCGA-COAD+READ, the precision--recall
curves differ more visibly among methods than the ROC curves, reflecting
the differences in AP. In TCGA-LUAD, the ROC and precision--recall curves overlap across methods,
and the precision--recall curves intersect.

\subsection{Comparison with Four-State ABMIL}

We further compared State-Aware with FourState-ABMIL to assess whether four-state supervision alone was sufficient to account for the observed performance (Table~\ref{tab:fourstate_abmil}).  In TCGA-COAD+READ (joint-positive prevalence 6.8\%), AP was 0.5566 versus
0.4965 with UNI2-h and 0.4410 versus 0.3859 with CONCH. In TCGA-LUAD
(joint-positive prevalence 8.6\%), AP was 0.2784 versus 0.2418 with UNI2-h
and 0.1659 versus 0.1529 with CONCH. AUROC was also higher for State-Aware in all four comparisons. This comparison indicates that four-state supervision alone did not reproduce the performance of State-Aware.

\begin{table}[htb]
\centering
\caption{Comparison of FourState-ABMIL and State-Aware Interaction MIL for
BRAF$^{+}$/MSI$^{+}$ prediction in TCGA-COAD+READ (prevalence 6.8\%) and
EGFR$^{+}$/TP53$^{+}$ prediction in TCGA-LUAD (prevalence 8.6\%) using
UNI2-h and CONCH representations.}
\label{tab:fourstate_abmil}
\small
\setlength{\tabcolsep}{4pt}
\renewcommand{\arraystretch}{1.05}
\begin{tabular}{lllcc}
\toprule
Cohort & Representation & Model & AUROC & AP \\
\midrule

TCGA-COAD+READ & UNI2-h & FourState-ABMIL
& 0.8978 $\pm$ 0.1122
& 0.4965 $\pm$ 0.2200 \\

TCGA-COAD+READ & UNI2-h & State-Aware
& \textbf{0.9080 $\pm$ 0.0965}
& \textbf{0.5566 $\pm$ 0.2015} \\

\addlinespace[2pt]

TCGA-COAD+READ & CONCH & FourState-ABMIL
& 0.8532 $\pm$ 0.0404
& 0.3859 $\pm$ 0.1075 \\

TCGA-COAD+READ & CONCH & State-Aware
& \textbf{0.8876 $\pm$ 0.0473}
& \textbf{0.4410 $\pm$ 0.0446} \\

\addlinespace[2pt]

TCGA-LUAD & UNI2-h & FourState-ABMIL
& 0.6542 $\pm$ 0.1565
& 0.2418 $\pm$ 0.1352 \\

TCGA-LUAD & UNI2-h & State-Aware
& \textbf{0.6725 $\pm$ 0.1746}
& \textbf{0.2784 $\pm$ 0.1714} \\

\addlinespace[2pt]

TCGA-LUAD & CONCH & FourState-ABMIL
& 0.5550 $\pm$ 0.1169
& 0.1529 $\pm$ 0.0603 \\

TCGA-LUAD & CONCH & State-Aware
& \textbf{0.5725 $\pm$ 0.1285}
& \textbf{0.1659 $\pm$ 0.1078} \\

\bottomrule
\end{tabular}
\end{table}

\subsection{Four-State Molecular Separation}
Overall differences across the four molecular states were assessed using the
Kruskal--Wallis test (Table~\ref{tab:state_separation}). Pairwise comparisons
between the joint-positive state and each alternative state used one-sided
Mann--Whitney $U$ tests, with Benjamini--Hochberg correction applied to the
three pairwise comparisons within each cohort.
\begin{table*}[!htb]
\caption{Comparison of out-of-fold joint-positive prediction scores across
the four molecular states for BRAF$^{+}$/MSI$^{+}$ in TCGA-COAD+READ and
EGFR$^{+}$/TP53$^{+}$ in TCGA-LUAD using State-Aware UNI2-h predictions,
including pairwise comparisons between the joint-positive state and each
alternative molecular state.}
\label{tab:state_separation}
\centering
\footnotesize
\setlength{\tabcolsep}{5pt}
\renewcommand{\arraystretch}{0.95}
\begin{tabular}{@{}lllccc@{}}
\toprule
\textbf{Cohort} &
\textbf{Joint-positive group} &
\textbf{Compared group} &
\textbf{Median $p_{\mathrm{joint}}$} &
\textbf{$P_{\mathrm{sup}}$} &
\textbf{BH-adjusted $p$} \\
\midrule
TCGA-COAD+READ &
BRAF$^{+}$/MSI$^{+}$ & BRAF$^{-}$/MSI$^{-}$ &
$0.2758$ vs.\ $0.0011$ & $0.943$ & $8.90\times10^{-16}$ \\
&
BRAF$^{+}$/MSI$^{+}$ & BRAF$^{-}$/MSI$^{+}$ &
$0.2758$ vs.\ $0.1628$ & $0.637$ & $0.040$ \\
&
BRAF$^{+}$/MSI$^{+}$ & BRAF$^{+}$/MSI$^{-}$ &
$0.2758$ vs.\ $0.0868$ & $0.682$ & $0.040$ \\
\midrule
TCGA-LUAD &
EGFR$^{+}$/TP53$^{+}$ & EGFR$^{-}$/TP53$^{-}$ &
$0.0651$ vs.\ $0.0307$ & $0.713$ & $6.30\times10^{-5}$ \\
&
EGFR$^{+}$/TP53$^{+}$ & EGFR$^{-}$/TP53$^{+}$ &
$0.0651$ vs.\ $0.0415$ & $0.653$ & $0.00248$ \\
&
EGFR$^{+}$/TP53$^{+}$ & EGFR$^{+}$/TP53$^{-}$ &
$0.0651$ vs.\ $0.0590$ & $0.559$ & $0.231$ \\
\bottomrule
\end{tabular}
\end{table*}

Here, $P_{\mathrm{sup}}$
denotes the probability of superiority, defined as the probability that a
randomly chosen joint-positive case receives a higher
$p_{\mathrm{joint}}$ score than a randomly chosen case from the comparison
state.

In TCGA-COAD+READ, prediction scores differed across states
(Kruskal--Wallis $H=96.19$, $p=1.02\times10^{-20}$), and the
BRAF$^{+}$/MSI$^{+}$ group had an elevated median score relative to the
double-negative and both single-positive groups. All three pairwise
comparisons remained significant after Benjamini--Hochberg correction,
supporting higher joint-positive scores than in the double-negative and both
single-positive states. In TCGA-LUAD, prediction scores also differed across
states (Kruskal--Wallis $H=20.75$, $p=1.19\times10^{-4}$). The
EGFR$^{+}$/TP53$^{+}$ group had an elevated median score relative to the
double-negative and TP53-only groups, and both comparisons remained
significant after correction. Joint-positive scores were not significantly
higher than those of the EGFR-only group, indicating overlap between these
two molecular configurations. The pattern of molecular-state separation
differed between TCGA-COAD+READ and TCGA-LUAD.

\subsection{Component Ablation}

Table~\ref{tab:ablation} evaluates the contributions of interaction modeling,
auxiliary biomarker supervision, and four-state molecular supervision. The no-interaction variant sets $i=0$, the no-auxiliary variant sets
$\lambda_{\mathrm{aux}}=0$, and the binary-supervision variant replaces the
four-state objective with binary joint-positive supervision while retaining
the biomarker-specific auxiliary losses.
Replacing four-state supervision with binary joint-positive supervision
reduced both AUROC and AP across both cohorts and both foundation-model
representations, indicating that retaining molecular-state information during
training improved joint-positive prediction. Removing the interaction module
reduced AP in both TCGA-COAD+READ experiments and in TCGA-LUAD with UNI2-h,
while the no-interaction variant achieved higher AP with CONCH in
TCGA-LUAD. Removing auxiliary biomarker supervision reduced AP in both
TCGA-COAD+READ experiments and with UNI2-h in TCGA-LUAD, while AP with
CONCH in TCGA-LUAD changed little. 

\begin{table*}[htb]
\caption{Component ablation of State-Aware Interaction MIL for
BRAF$^{+}$/MSI$^{+}$ prediction in TCGA-COAD+READ (prevalence 6.8\%) and
EGFR$^{+}$/TP53$^{+}$ prediction in TCGA-LUAD (prevalence 8.6\%) using
UNI2-h and CONCH representations.}
\label{tab:ablation}
\centering
\setlength{\tabcolsep}{3pt}
\renewcommand{\arraystretch}{0.95}

\resizebox{\textwidth}{!}{%
\begin{tabular}{@{}llcccc@{}}
\toprule
\textbf{Cohort} &
\textbf{Variant} &
\multicolumn{2}{c}{\textbf{UNI2-h}} &
\multicolumn{2}{c}{\textbf{CONCH}} \\
\cmidrule(lr){3-4}
\cmidrule(lr){5-6}
& &
\textbf{AUROC} & \textbf{AP} &
\textbf{AUROC} & \textbf{AP} \\
\midrule

TCGA-COAD+READ
& \textbf{Full State-Aware}
& $\mathbf{0.9080\pm0.0965}$ & $\mathbf{0.5566\pm0.2015}$
& $\mathbf{0.8876\pm0.0473}$ & $\mathbf{0.4410\pm0.0446}$ \\

&
No interaction
& $0.9038\pm0.0927$ & $0.5178\pm0.1568$
& $0.8281\pm0.0703$ & $0.3039\pm0.0993$ \\

&
No auxiliary supervision
& $0.9009\pm0.1002$ & $0.4604\pm0.1346$
& $0.8611\pm0.0516$ & $0.3604\pm0.0900$ \\

&
Binary joint supervision
& $0.8786\pm0.0810$ & $0.4542\pm0.1835$
& $0.8546\pm0.0594$ & $0.3920\pm0.1024$ \\

\midrule

TCGA-LUAD
& \textbf{Full State-Aware}
& $0.6725\pm0.1746$ & $\mathbf{0.2784\pm0.1714}$
& $0.5725\pm0.1285$ & $0.1659\pm0.1078$ \\

&
No interaction
& $\mathbf{0.7107\pm0.1035}$ & $0.2604\pm0.1296$
& $0.5561\pm0.1510$ & $\mathbf{0.1882\pm0.1128}$ \\

&
No auxiliary supervision
& $0.6467\pm0.1849$ & $0.2553\pm0.1704$
& $\mathbf{0.5923\pm0.1406}$ & $0.1665\pm0.0895$ \\

&
Binary joint supervision
& $0.6455\pm0.0969$ & $0.2079\pm0.1613$
& $0.4872\pm0.1093$ & $0.1063\pm0.0332$ \\

\bottomrule
\end{tabular}%
}
\end{table*}

\subsection{Spatial Attention Analysis}
\begin{figure*}[htb]
    \centering
    \includegraphics[width=\textwidth]{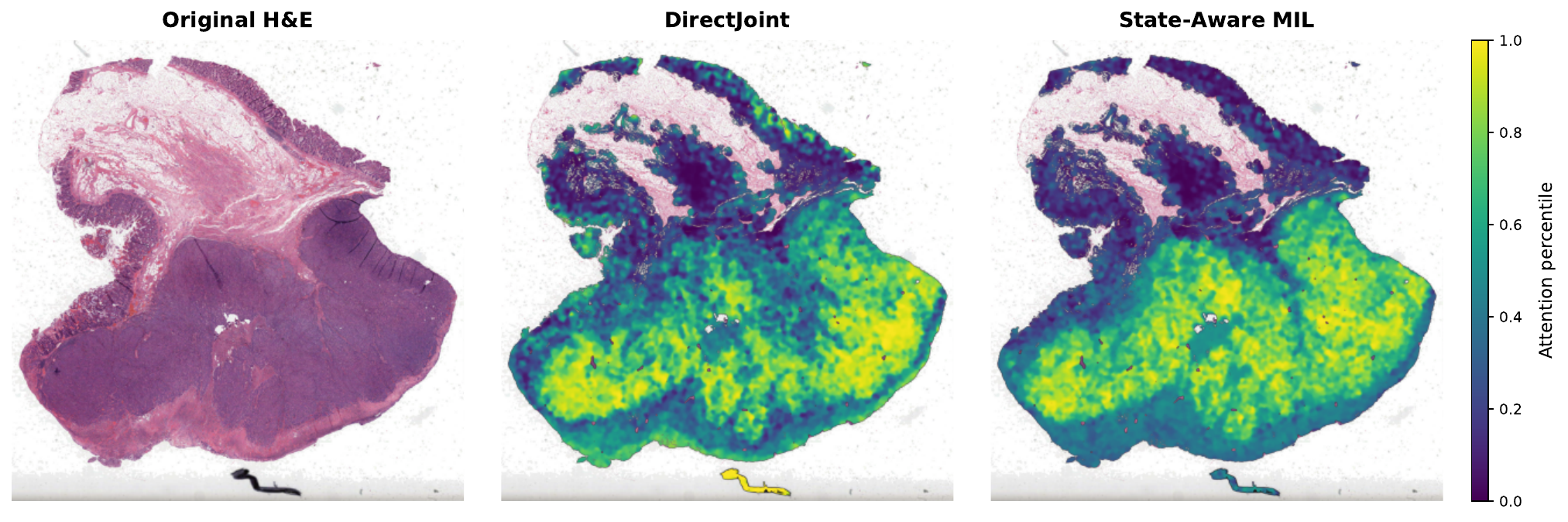}
\caption{Spatial attention patterns for BRAF/MSI joint phenotype prediction
in a held-out TCGA-COAD+READ case using DirectJoint and State-Aware
Interaction MIL.}
    \label{fig:crc_attention}
\end{figure*}

\begin{figure*}[htb]
    \centering
    \includegraphics[width=\textwidth]{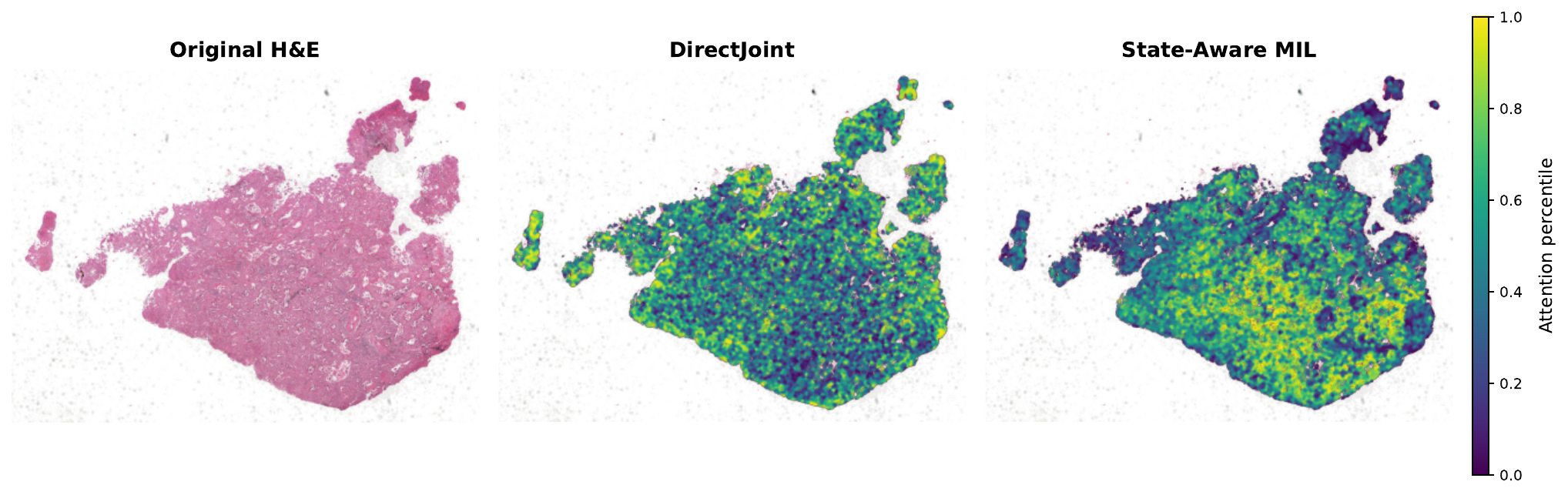}
\caption{Spatial attention patterns for EGFR/TP53 joint phenotype prediction
in a held-out TCGA-LUAD case using DirectJoint and State-Aware Interaction
MIL.}
    \label{fig:luad_attention}
\end{figure*}

Attention visualizations for held-out TCGA-COAD+READ and TCGA-LUAD cases
are shown in Figs.~\ref{fig:crc_attention} and~\ref{fig:luad_attention},
respectively, for DirectJoint and State-Aware Interaction MIL.

For State-Aware, joint attention is computed from the geometric mean of the
two biomarker-specific attention weights,
$\sqrt{\alpha_{A,i}\alpha_{B,i}}$, followed by normalization across tissue
tiles. Percentile scaling and spatial smoothing are applied only for
visualization. In TCGA-COAD+READ, both models
focus on overlapping tumor-containing regions, while State-Aware
concentrates attention within more localized areas of the tissue. These
regions may contain morphological patterns associated with the joint
BRAF/MSI phenotype. In TCGA-LUAD, DirectJoint distributes attention across
a wider portion of the slide, while State-Aware concentrates attention
within a localized tumor-bearing region. This difference may reflect
attention to localized histological patterns associated with the joint
EGFR/TP53 phenotype.

\section{Discussion}
The performance pattern differed between AP and AUROC: improvements in AP were consistent across both foundation-model representations, whereas the AUROC pattern depended on the representation. Given joint-positive prevalences of 6.8\% in TCGA-COAD+READ and 8.6\% in
TCGA-LUAD, AP provides a useful measure of precision--recall performance,
with these prevalences providing reference levels for interpretation. Joint-positive prediction performance was lower in TCGA-LUAD
than in TCGA-COAD+READ for all methods with both UNI2-h and CONCH
representations. The four-state analysis showed different separation patterns across the two
cohorts: BRAF$^{+}$/MSI$^{+}$ scores differed from all three alternative
states in TCGA-COAD+READ, whereas EGFR$^{+}$/TP53$^{+}$ scores overlapped
with the EGFR-only state in TCGA-LUAD. 

The consistent reduction under binary supervision suggests that retaining the non-joint molecular configurations provides useful supervisory information beyond a binary joint endpoint. Binary joint prediction combines the double-negative and two single-positive configurations into a single negative class, whereas four-state supervision preserves these molecular distinctions during training. The four-state score analysis
supports this interpretation because joint-positive prediction scores
differed across the molecular configurations. The effects of interaction modeling and auxiliary biomarker supervision
varied across cohorts and foundation-model representations: both improved AP
in the two TCGA-COAD+READ experiments and with UNI2-h in TCGA-LUAD,
whereas with CONCH in TCGA-LUAD, removing the interaction module increased
AP and removing auxiliary supervision produced little change. The attention analysis shows that the prediction formulation can affect the
spatial distribution of model attention in addition to the final slide-level
prediction. Biomarker-specific representations and their interaction may influence which
tissue regions receive greater attention during joint phenotype prediction.

\paragraph{Limitations.}
The evaluation is limited to retrospective TCGA cohorts and does not include
independent external validation across institutions or patient populations.
The experiments are restricted to two cancer--biomarker pairs and two frozen
pathology foundation-model representations. The limited number of
joint-positive cases within individual test folds contributes to variability
in the performance estimates.

\section{Conclusion}

We presented State-Aware Interaction MIL for rare joint molecular phenotype
prediction from whole-slide histology. The method preserves
biomarker-specific representations, models their interaction, and supervises
the complete four-state molecular configuration. Across TCGA-COAD+READ
(6.8\% joint-positive) and TCGA-LUAD (8.6\% joint-positive), State-Aware
improved AP with both UNI2-h and CONCH representations, while improvement
in AUROC was observed with CONCH. The ablation experiments showed that
retaining the complete molecular configuration during training improved
joint-positive prediction, while the four-state analysis showed that
prediction scores differed across molecular states. These results support
structured molecular-state supervision for predicting rare co-occurring
biomarker phenotypes from pathology foundation-model representations.

\section*{Code Availability}

The implementation of State-Aware Interaction MIL, including model training,
evaluation, baseline implementations, and analysis scripts, is publicly available at
\url{https://github.com/raajuuu1998/StateAwareMIL}.

\end{document}